\documentclass{article}

\usepackage{arxiv}

\usepackage{times}

\usepackage{soul}
\usepackage{url}
\usepackage[hidelinks]{hyperref}
\usepackage[utf8]{inputenc}
\usepackage[small]{caption}
\usepackage{graphicx}
\usepackage{amsmath}
\usepackage{booktabs}
\usepackage{tabularx}
\usepackage{listings}
\usepackage{tcolorbox}
\usepackage[ruled,vlined]{algorithm2e}
\usepackage{nicefrac}       
\usepackage{microtype}      
\usepackage{amsthm}
\usepackage{amsfonts}

\theoremstyle{definition}

\title{SAG-VAE: End-to-end Joint Inference of Data Representations and Feature Relations}

\author{
  Chen Wang \\
  Department of Computer Science\\
  Rutgers University - New Brunswick\\
  Piscataway, NJ 08854, USA \\
  \texttt{chen.wang.cs@rutgers.edu} \\
  \And
 Chengyuan Deng \\
  Department of Computer Science\\
  Rutgers University - New Brunswick\\
  Piscataway, NJ 08854, USA \\
  \texttt{charles.deng@rutgers.edu} \\
    \AND
  Vladimir Ivanov \\
  Department of Computer Science\\
  Rutgers University\\
  Piscataway, NJ 08854, USA \\
  \texttt{vladimir.ivanov@rutgers.edu} \\
}

\begin{document}
\maketitle

\begin{abstract}
The ability to capture relations within data can provide the much needed inductive bias for robust and interpretable Machine Learning algorithms. Variational Autoencoder (VAE) is a promising candidate for such purpose thanks to their power in data representation inference, but its vanilla form and common variations cannot process feature relations. In this paper, inspired by recent advances in relational learning with graph neural networks, we propose the \emph{S}elf-\emph{A}ttention \emph{G}raph \emph{V}ariational \emph{A}uto\emph{E}ncoder (SAG-VAE) model which can simultaneously learn feature relations and data representations in an end-to-end manner. The SAG-VAE is trained by jointly inferring the posterior distribution of two types of latent variables, which respectively represent the data and the feature relations. The feature relations are represented as a graph structure, and the presence of each edge is determined by a Gumbel-Softmax distribution. The generative model is accordingly parameterized by a graph neural network with a special attention mechanism we introduced in the paper. Therefore, the SAG-VAE model can generate via graph convolution and be trained via backpropagation. Experiments based on graphs show that SAG-VAE is capable of approximately retrieving edges and links between vertices based entirely on feature observations. Furthermore, experiments on image data illustrate that the learned feature relations can provide the SAG-VAE robustness against perturbations in image reconstruction and sampling. The learned feature relations as graph adjacency matrices are observed to be structured, which provides intuitive interpretability of the models.
\end{abstract}


\section{Introduction}
\label{sec:introduction}
In practice, data often comes with complex relations between features which are not explicitly visible, and extracting this structural information has been a crucial, yet challenging, task in the field of Machine Learning. Recently, renewed interest in relational and structure learning has been largely driven by the development of new end-to-end Neural Network and Deep Learning frameworks \cite{bordes2013translating,xu2018representation,van2018relational}, with multiple promising results reported. This renewed drive in relational structure inference using Neural Networks can be partially attributed to current efforts to overcome the limited generalization capabilities of Deep Learning \cite{battaglia2018relational}. More importantly, learning the relational structure with Neural Network models has several inherent advantages:  strong and efficient parameterization ability of Deep Learning can extract essential relational information and perform large-scale inference, which are considered difficult with other learning algorithms.\par

Recently, research in relational learning using Neural Networks has largely focused on sequential generation/prediction of dynamical systems, while static data has been largely ignored \cite{louizos2017causal,kipf2018neural,sanchez2018graph}.  At their core, these algorithms use either one or a combination of Graph Neural Networks (GNNs) \cite{scarselli2008graph,kipf2016semi,wu2019comprehensive} and Variational Autoencoders (VAEs) \cite{kingma2013auto}. The former provide a convenient framework for relational operations through the use of graph convolutions \cite{schlichtkrull2018modeling}, and the latter offer a powerful Bayesian inference method to learn the distribution of the latent graph structure of data. Inspired by these recently developed methods, we devised a Neural Network based algorithm for relational learning on graph data.\par

In this paper, inspired by the recent advances in the field of GNNs and VAEs, we propose Self-Attention Graph Variational Autoencoder (SAG-VAE), a novel VAE framework that jointly learns data representation and latent structure in an end-to-end manner. SAG-VAE utilizes the gumbel-softmax reparameterization \cite{jang2016categorical} to infer the graph adjacency matrix, and employs a novel Graph Convolutional Network (also proposed by this paper) as the generative network. During the generative process, a sampled adjacency matrix will serve as the graph edge information for the novel Graph Network, and a sampled data representation will be fed into the network to generate data. Based on this framework, SAG-VAE will be able to directly infer the posterior distributions of both the data representation and relational matrix based simply on gradient descent. \par

Several experiments are carried out with multiple data sets of different kinds to test the performances. We observe in the experiments that SAG-VAE can learn organized latent structures for homogeneous image data, and the interpretation can match the nature of the certain type of image. Also, for graph data with known connections, SAG-VAE can retrieve a significant portion of the connections based entirely on feature observations. Based on these performances, we argue that SAG-VAE can serve as a general relational structure learning method from data. Furthermore, since SAG-VAE is a general framework compatible with most Variational Autoencoders, it is straightforward to combine advanced VAEs with SAG-VAE to create more powerful models. \par

The rest of the paper is arranged as follows: Section \ref{sec:relatedwork} conducts a literature review regarding methods related to the paper; Section \ref{sec:methodology} introduces the background and discuss the proposed SAG-VAE; Experimental results are shown in section \ref{sec:experiment}, and the implications are discussed; And finally, section \ref{sec:conclusion} gives a general conclusion of the paper. \par

\section{Related Work}
\label{sec:relatedwork}
Early interest in learning latent feature relations and structures partly stems from questions over causality in different domains \cite{granger1969investigating,kuipers1984causal}. Before the era of machine learning, methods in this field substantially relied on domain knowledge and statistical scores \cite{brillinger1976identification,watts1998collective}, and most of them work only on small-scale problems. The recent advancements in machine learning prompted the development of large-scale and trainable models on learning feature relations \cite{linderman2016bayesian,yang2018glomo}. However, most of the aforementioned methods are domain-specified and are usually not compatible with general-purpose data. Thus, these models are not quantitatively evaluated or compared in this paper. \par
Feature relational learning in Neural Networks can trace its history from sequential models. Recurrent Neural Network (RNN) and its variants like LSTM \cite{hochreiter1997long} are the early examples of relational learning methods, although their aspect of `feature relation' has been overwhelmed by their success in sequential modeling. After the emerge of Deep Learning, researchers in the domain of Natural Language Processing first built `neural relational' models to exploit the relations between features \cite{mikolov2013efficient}. Recently, a variety of notable methods on neural relational learning, such as AIR \cite{eslami2016attend}, (N-)REM \cite{greff2017neural,van2018relational} and JK network \cite{xu2018representation}, have achieved state-of-the-art performances by adopting explicit modeling of certain relations. However, although the models discussed above are powerful, most of them assume a known relational structure given by the data or experts, which means they do \emph{not} have the ability in \emph{extracting} feature relations. \par

More recently, the idea of leveraging graph neural networks to learn feature relations has grasped considerable interests \cite{battaglia2018relational}. The graph neural network model was originally developed in the early 2000s \cite{sperduti1997supervised,gori2005new,scarselli2008graph}, and it has been intensively improved by a series of research efforts \cite{bruna2013spectral,defferrard2016convolutional,henaff2015deep}. And finally, \cite{kipf2016semi} proposed the well-renowned Graph Convolutional Network (GCN) model which established the framework of modern graph neural networks. Graph neural networks are increasingly popular in the exploration and exploitation of feature relations \cite{sanchez2018graph,schlichtkrull2018modeling, Wang2018NerveNetLS}, and there are several methods in this domain similar to the proposed SAG-VAE. For instance, \cite{kipf2018neural} embeds a graph neural network into the framework of Variational Autoencoders to learn the latent structure for dynamic models, and \cite{Grover2019GraphiteIG} designs an iterative refining algorithm to extract the graph structure. Furthermore, \cite{velivckovic2017graph} proposed the Variational Graph Autoencoder (VGAE) that can reconstruct graph edges from feature observations and limited number of given edges. The VGAE model provides a strong baseline to evaluate graph edge retrieval. 
Apart from the graph neural networks, this paper is also closely related to Variational Autoencoders (VAEs) \cite{kingma2013auto} and the Gumbel-Softmax distribution \cite{jang2016categorical}. Among the numerous variations of the VAEs, \cite{kipf2016variational} devises an auto-encoding inference structure composed by a Graph Convolutional Network-based encoder and an inner product-based decoder, which can accomplish tasks similar to the SAG-VAE. Furthermore, \cite{pan2018adversarially} elaborated on the idea to use VAEs to learn explicit graph structure. Gumbel-softmax was introduced by \cite{jang2016categorical} to provide a `nearly-discrete' distribution compatible with reparametrization and backpropagation. Based on this technique, we can compute gradients for each edge, which is considered impossible with the categorical distribution.

\section{Method}
\label{sec:methodology}
\subsection{Background}
\subsubsection{Graph Convolution Networks}
We first introduce Graph Convolutional Networks following the framework of \cite{kipf2016semi}. A graph is denoted as $\boldsymbol{G} = (\boldsymbol{V},\boldsymbol{E})$, where $\boldsymbol{V}$ is the set of vertices and $\boldsymbol{E}$ is the set of edges. The vertices and their features are denoted by a $n\times d$ matrix, where $n=|\boldsymbol{V}|$ and $d$ is number of features.  A graph adjacency matrix $\boldsymbol{A}$ of size $n\times n$ is adopted to indicate the edge connections, and $\boldsymbol{\hat{A}}=\boldsymbol{A}+\boldsymbol{I}$ is used to introduce relevance for each vertex itself. A feed-forward layer is characterized by the following equation:
\begin{equation}
\label{equ:graphaggre}
\begin{aligned}
    \boldsymbol{H}^{(l+1)} &= f_{\boldsymbol{W}}(\boldsymbol{H}^{(l)},\boldsymbol{A}) \\
            &= \sigma(\boldsymbol{\hat{D}}^{-\frac{1}{2}} \boldsymbol{\hat{A}} \boldsymbol{\hat{D}}^{-\frac{1}{2}} \boldsymbol{H}^{(l)} \boldsymbol{W}^{(l)} )\\
            &=\sigma(\boldsymbol{\tilde{A}}\boldsymbol{H}^{(l)} \boldsymbol{W}^{(l)})
\end{aligned}
\end{equation}
where $\boldsymbol{\hat{D}}$ is the diagonal matrix with $\boldsymbol{\hat{D}}_{s,s}=\sum_{t}\boldsymbol{\hat{A}}_{s,t}$, and $\boldsymbol{\tilde{A}}=\boldsymbol{\hat{D}}^{-\frac{1}{2}}(\boldsymbol{A}+\boldsymbol{I})\boldsymbol{\hat{D}}^{-\frac{1}{2}}$ is the normalized adjacency matrix.
\subsubsection{Variational Autoencoders}
Variational Autoencoders (VAEs) have been witnessed to be one of the most efficient approaches to infer latent representations of the data \cite{kingma2013auto}. Following the standard notation, we use $p(\cdot)$ to denote the real distribution and $q(\cdot)$ for the variational distribution. Therefore, the inference model $q(\boldsymbol{Z}|\boldsymbol{X})$ and the generative model $p(\boldsymbol{X}|\boldsymbol{Z})$ as:
\begin{equation}
\begin{aligned}
    q(\boldsymbol{Z}|\boldsymbol{X}) &= \prod_{i=1}^{m} q_{\phi}(\boldsymbol{z}_i|\boldsymbol{x}_{i})\\
    p(\boldsymbol{X}|\boldsymbol{Z}) &= \prod_{i=1}^{m} p_{\theta}(\boldsymbol{x}_{i}|\boldsymbol{z}_i)
\end{aligned}
\end{equation}
Where $m$ stands for the amount of data. And under the Gaussian prior used in the original paper \cite{kingma2013auto}, the inference network will be:
\begin{equation}
q_{\phi}(\boldsymbol{z}_i|\boldsymbol{x}_{i}) = \mathcal{N} (\boldsymbol{z}_i| \mu_{\phi(\boldsymbol{x}_{i})}, \texttt{diag}(\sigma^{2}_{\phi(\boldsymbol{x}_{i})}))
\end{equation}
and the optimization objective was given as the format of Evidence Lower Bound (ELBO): 
\begin{equation}
\begin{aligned}
     \log p(\boldsymbol{X}) \geq & -\mathcal{L}(\theta, \phi) \\
     = & \mathbb{E}_{\boldsymbol{Z} \sim q_{\phi}(\boldsymbol{Z}|\boldsymbol{X})}[\log p_{\theta}(\boldsymbol{X}|\boldsymbol{Z})] \\
      &\hspace{1.0cm} -D_{KL}(q_{\phi}(\boldsymbol{Z}|\boldsymbol{X})||p(\boldsymbol{Z}))
\end{aligned}
\end{equation}
For conjugate priors like Gaussian, the KL-divergence can be computed analytically to avoid the noise in Monte-Carlo simulations.
\subsubsection{Gumbel-Softmax Distribution}
To introduce feature relations as a graph structure, the most straightforward approach is to represent each edge connection as a Bernoulli random variable. Alas, there is no properly-defined gradients for discrete distributions like Bernoulli. Hence, to train the model in a backpropagation fashion, we must have some alternatives to the truly discrete distribution. Thanks to recent advances in Bayesian Deep Learning, we are able to utilize Gumbel-Softmax distribution \cite{jang2016categorical} to simulate Bernoulli/categorical distributions. A simplex-valued random variable $\boldsymbol{a}$ from a Gumbel-Softmax distribution is a $k$-length vector characterized by the follows:
\begin{equation}
\label{equ:gumbel-dist}
\boldsymbol{a}_{1:k} = \big(\frac{\exp((\log(\alpha_{k})+G_{k})/\tau)}{\sum_{k=1}^{K}\exp((\log(\alpha_{k})+G_{k})/\tau)}\big)_{1:k}
\end{equation}
where $\alpha_{k}$ is proportional to the Bernoulli/categorical probability and $G_{k}$ is a noise from the Gumbel distribution. The subscript $1:k$ indicates a softmax vector, and $\tau$ is the temperature variable that control the `sharpness' of the distribution. A higher $\tau$ will make the distribution closer to a uniform one, and a lower $\tau$ will lead to a more discrete-like distribution. \par
Notice that the above equation is not a density function: the density function for the Gumbel-Softmax distribution is complex, and we usually do not use it in practice. What is of our interest is that we can design neural networks to learn $\log(\alpha_{k})$ for each class (in the case of graph edge connection, the number of classes is 2 since we want to approximate Bernoulli), and although the output of the neural network is not necessarily valid distributions, we can apply the reparametrization trick and the transformation of equation \ref{equ:gumbel-dist} to get simplex-valued vectors. In this way, the neural network to learn $\log(\alpha_{k})$ (encoding network) can be trained by backpropagation since the gradients of equation \ref{equ:gumbel-dist} is well-defined.
\subsection{Inference of SAG-VAE}
Based on the above strategy, we introduce another latent variable $\boldsymbol{A}$, which represents the distribution of the adjacency matrix of the graph. We only consider undirected graph in this paper, so the distribution an be factorized into $p(\boldsymbol{A})=\prod_{s=1}^{n}\prod_{t=s+1}^{n}p(\boldsymbol{A}_{s,t})$. Following the doctrine of variational inference, we use the Gumbel-Softmax distribution to approximate the probability for each edge:
\begin{equation}
\label{equ:gumbelapprox}
    q_{\phi}(\boldsymbol{A}_{s,t}|\boldsymbol{X}) = \texttt{Gumbel-Softmax}(\phi_{1}(\boldsymbol{A}_{s,t}|\boldsymbol{X}))
\end{equation}
\par
Notice that in equation \ref{equ:gumbelapprox} there is no index on $\boldsymbol{X}$, which means the learned adjacency matrix is a shared structure (amortized inference) and should be averaging over the input. In practice, one can apply Gumbel-Softmax to each $\phi_{1}(\boldsymbol{A}_{s,s}|\boldsymbol{X}_{i})$, and averaging over the probability:
\begin{equation}
\label{equ:gumbelapproxave}
    q_{\phi}(\boldsymbol{A}_{s,t}|\boldsymbol{X}) = \frac{1}{m}\sum_{i=1}^{m}\texttt{Gumbel-Softmax}(\phi_{1}(\boldsymbol{A}_{s,t}|\boldsymbol{X}_{i}))
\end{equation}
as this will make the estimation of the KL-divergence part more robust. We will discuss more on this issue further in the later paragraphs.
\begin{figure}[!h]
\centering
\includegraphics[width=0.7\textwidth]{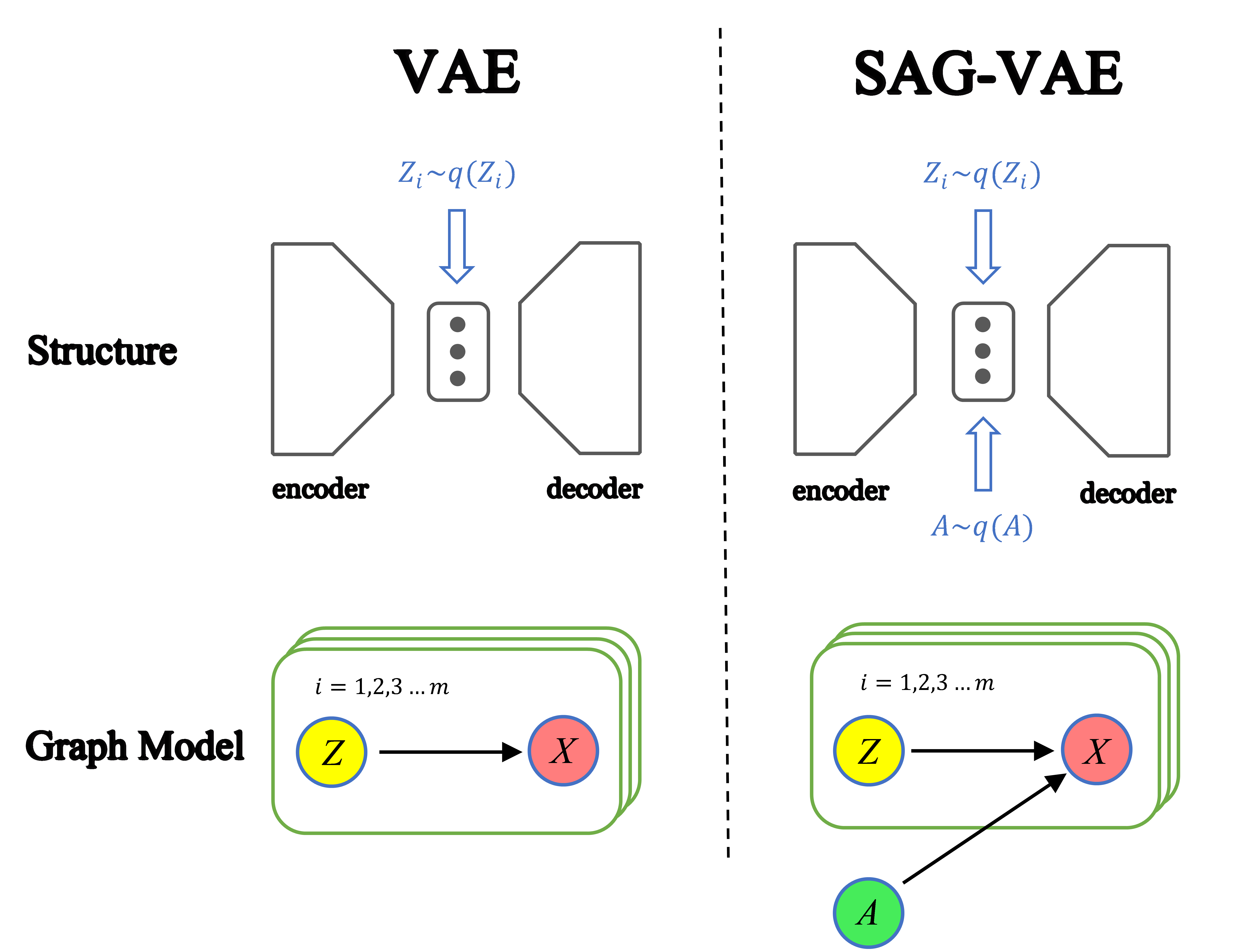}
\caption{\label{fig:modelstructure}The model structure and graphical model of SAG-VAE.}
\end{figure}
\par
Taking back the original $\boldsymbol{Z}$ variable, the joint posterior is $p(\boldsymbol{Z},\boldsymbol{A}|\boldsymbol{X})$. Figure \ref{fig:modelstructure} illustrates the difference between the vanilla VAE and the SAG-VAE. Observing from the graphical model of SAG-VAE, since $\boldsymbol{A}$ and $\boldsymbol{Z}$ are considered not $d$-separated, they are not necessarily independent given $\boldsymbol{X}$. Nevertheless, to simplify computation, we perform the conditional independence approximation on the variational distributions:\\
\begin{equation}
\label{equ:posteriorapprox}
p(\boldsymbol{Z}, \boldsymbol{A}|\boldsymbol{X}) \approx q_{\phi_1}(\boldsymbol{Z}|\boldsymbol{X})q_{\phi_2}(\boldsymbol{A}|\boldsymbol{X})
\end{equation}
Crucially, equation \ref{equ:posteriorapprox} allows the posterior distributions to be separated, and therefore avoids noisy and expensive Monte-Carlo simulation of the joint KL-divergence. With the similar derivation developed in \cite{kingma2013auto}, one can get the new ELBO of our model:\\
\begin{equation}
\label{equ:inferenceTarget}
\begin{aligned}
     \log p(\boldsymbol{X}) &\geq -\mathcal{L}(\theta, \phi_1, \phi_2)\\
     &= \mathbb{E}_{\boldsymbol{Z}\sim q_{\phi_1}(\boldsymbol{Z}|\boldsymbol{X}),\boldsymbol{A}\sim q_{\phi_2}(\boldsymbol{A}|\boldsymbol{X})}[\log p_{\theta}(\boldsymbol{X}|\boldsymbol{Z},\boldsymbol{A})]
     \\&\hspace{1.0cm} -D_{KL}[q_{\phi_1}(\boldsymbol{Z}|\boldsymbol{X})||p(\boldsymbol{Z})]
     \\&\hspace{1.0cm} -D_{KL}[q_{\phi_2}(\boldsymbol{A}|\boldsymbol{X})||p(\boldsymbol{A})]
\end{aligned}
\end{equation}
The posterior distribution of $\boldsymbol{Z}$ is characterized by a learned Gaussian distribution, and the prior $p(\boldsymbol{Z})$ is standard Gaussian. We omit more complicated priors developed recently since our focus is not on powerful data representation. The posterior distribution of $\boldsymbol{A}$ is ccharacterized by the learned Gumbel-Softmax distribution, and the prior of $\boldsymbol{A}$ is a Bernoulli distribution with one-hot, uniform or specified values. \par
For SAG-VAE, we need the dimension of the hidden representation to be equal to the number of dimension (one can see the reason in section \ref{subsec:sagn}). Therefore, we propose two types of implementations. The first one is to apply a set of hidden distributions for each data point, as it is usually applied in ordinary VAEs; and the second one is to learn a distribution for each dimension. Noticeably, the latter scheme will lead to high-quality reconstruction results, albeit with the cost that the model becomes more vulnerable to noise/perturbations and sampling from the SAG-VAE becomes difficult. Nevertheless, the advantages of robustness and noise-resistance of SAG-VAE are more obvious with the second implementation. \par
Another issue to notice is the computation of the KL-divergence term $D_{KL}[q_{\phi_2}(\boldsymbol{A}|\boldsymbol{X})||p(\boldsymbol{A})]$. Notice that for the SAG-VAE with data point-wise representation, with the implementation based on equation \ref{equ:gumbelapproxave}, the KL divergence will become:
$$\frac{1}{m}\sum_{i=1}^{m}\sum_{j=1}^{n^{2}-n}D_{KL}(q_{\phi_2}(\boldsymbol{A}_{j=\{s,t\}}|\boldsymbol{X}_{i})||p(\boldsymbol{A}_{j=\{s,t\}}))$$
This function is not properly normalized as the summation depends on $n$ but there is no such parameter on the denominator. For the per-dimension version of SAG-VAE, although we do have an additional $\frac{1}{n}$ factor, this KL-divergence term can still be way too dominating as the summation is of $O(n^{2})$ terms. Thus, inspired by the idea in \cite{chou2019generated}, we use a $\beta_{A}=\frac{1}{n^{2}-n}$ to normalize the KL-divergence term ($\beta_{A}D_{KL}[q_{\phi_2}(\boldsymbol{A}|\boldsymbol{X})||p(\boldsymbol{A})]$) and improve the performance.
\subsection{Self-attention Graph Generative Network}
\label{subsec:sagn}
The generative network of SAG-VAE is composed by a novel Self-attention Graph Neural Network model design in this paper. We can denote this in a short-handed notation:
\begin{equation}
\label{equ:gennetwork}
p_{\theta}(\boldsymbol{X}_{i}|\boldsymbol{Z}_{i},\boldsymbol{A}) = \textbf{SA-GNN}_{\theta}(\boldsymbol{Z}_{i}, \boldsymbol{A})
\end{equation}
The Self-attention Graph Network follows the framework of \cite{kipf2016semi} for information aggregation. On the top of that, one significant difference is the introduction of the self-attention layer. The approach is similar to the mechanism in Self-attention GANs \cite{zhang2018self}, but instead of performing global attention regardless of geometric information, the self-attention layer in our model is based on the neighbouring nodes. The reason for adopting such a paradigm in this model is that the node features and edge connections are learned instead of given. And if a global unconditional attention is performed, the errors on the initialization stage will be augmented. \par
Suppose feature $\boldsymbol{H}^{(l)}$ is the output of the previous layer has the shape $[n \times d^{(l)}]$, where $n$ and $d^{(l)}$ represent the number of dimensions (vertices) and graph features, respectively. Now for node $i$ and any other node $j \in \mathcal{N}_{i}$ (neighboring vertices), the relevance value $e_{i,j}$ is computed as follows:
\begin{equation}
\label{equ:attentionRelevance}
e_{i,j} = (\boldsymbol{H}^{(l)}_{i}\boldsymbol{W_{l}})(\boldsymbol{H}^{(l)}_{j}\boldsymbol{W_{r}})^{T}
\end{equation}
where $\boldsymbol{W_{l}}$ and $\boldsymbol{W_{r}}$ are the $[d^{(l)} \times \bar{d}]$ convolution matrices to transform the $d$-dimension features to $\bar{d}$-dim attention features. Finally, having taken into consideration the graph edge connections as geometric information, we perform the softmax operation on the neighboring nodes of $i$ (including itself). Formally, the attention value will be computed as:\\
\begin{equation}
\label{equ:attentionSoftmax}
\alpha_{i,j} = \frac{\exp(e_{i,j})}{\sum_{j\in \mathcal{N}_{i}\cup\{i\}}\exp(e_{i,j})}, \forall j \in \mathcal{N}_{i}\cup\{i\}
\end{equation}
In practice, the above operation can be done before normalization in parallel by multiplying the relevance information computed by equation \ref{equ:attentionRelevance} with the adjacency matrix. This attention mechanism is similar to Graph Attention Network (GAT) \cite{velivckovic2017graph}, with one main difference being that in GAT the relevance features are aggregated and multiplied by a learnable vector, while in SA-GNN the relevance features are directly processed by dot products. After computing $\alpha_{i,j}$ for each pair and obtaining the matrix $\boldsymbol{\alpha}$, the attention result can be directly computed by matrix multiplication in the same manner of \cite{zhang2018self}:\\
\begin{equation}
\label{equ:attentionmul}
\boldsymbol{\bar{H}}^{(l)} = [\boldsymbol{\alpha}(\boldsymbol{H}^{(l)}\boldsymbol{W_{g}})]\boldsymbol{W_{f}}
\end{equation}
where $\boldsymbol{W_{g}}$ and $\boldsymbol{W_{f}}$ are the $[d^{(l)} \times \bar{d}]$ and $[\bar{d} \times d^{(l)}]$ transformation matrices, respectively. The main purpose of using the two matrix is to reduce computational cost. \par
To introduce more flexibility, we considered incorporating edge weights into the attention mechanism. The weights can be computed by the encoding matrix with a share structure of $q_{\phi_2}(\boldsymbol{A})$ network. Formally, this can be expressed as:\\
\begin{equation}
\label{equ:attentionWeights}
\boldsymbol{V} = \boldsymbol{\hat{\phi}_2}(\boldsymbol{X})
\end{equation}
where $\boldsymbol{\hat{\phi}_2}(\cdot)$ indicates a network shares the structure with $\boldsymbol{\phi_2}(\cdot)$ except the last layer. Meanwhile, the main diagonal of $\boldsymbol{V}$ will be set to 1. Therefore, equation \ref{equ:attentionSoftmax} can be revised into:\\
\begin{equation}
\label{equ:attentionSoftmaxWeighted}
\alpha_{i,j} = \frac{\exp(e_{i,j})V_{i,j}}{\sum_{j\in \mathcal{N}_{i}\cup\{i\}}\exp(e_{i,j})V_{i,j}}, \forall j \in \mathcal{N}_{i}\cup\{i\}
\end{equation}
And in a similar idea to \cite{zhang2018self}, the attention-based feature will be multiplied by a $\lambda$ coeffcient originally set as 0 and added to the features updated by the rules in vanilla GCN:\\
\begin{equation}
\label{equ:attentionOutput}
\boldsymbol{H}^{(l+1)}  = (\lambda\boldsymbol{\bar{H}}^{(l)} + \boldsymbol{\tilde{A}} \boldsymbol{H}^{(l)})\boldsymbol{W}^{(l)}
\end{equation}
where $\boldsymbol{W}^{(l)}$ is the convolution weights of the $l$-th layer. Based on the above equation, the network will first focus on learning the graph geometry (edges), and then using the attention mechanism to improve the generation quality. \par 
One potential issue of training VAEs is the so-called `posterior collapse', i.e., the posterior distribution becomes irrelevant from data when the decoder (generative model) is powerful. Graph neural networks are powerful models, so to make sure the posterior distributions are properly trained, we introduced the idea in \cite{dieng2018avoiding} to enforce correlation between the generated graph adjacency matrix and the output of each layer. Specifically, we use skip connection to interpolate the latent representation of data with the self-attention-processed information at each layer. This can be denoted as:
\begin{equation}
\boldsymbol{H}^{(l+1)} = \sigma(\lambda\boldsymbol{\bar{H}}^{(l)} + \boldsymbol{\tilde{A}} \boldsymbol{H}^{(l)})\boldsymbol{W}^{(l)} + \boldsymbol{\tilde{A}}\boldsymbol{H}^{(1)}\boldsymbol{\hat{W}}^{(l)}
\end{equation}
where $\sigma(\cdot)$ is the non-linear activation, $\boldsymbol{H}^{(1)}$ is the latent representation of the data (directly from $\boldsymbol{Z}$), and $\boldsymbol{\hat{W}}^{(l)}$ is the convolutional weight between the latent representation and the current layer. For the last layer, we apply activation after amalgamating the information:
\begin{equation}
\label{equ:finalOutput}
\boldsymbol{H}^{(L)}  = \sigma \big((\lambda\boldsymbol{\bar{H}}^{(L-1)}  + \boldsymbol{\tilde{A}} \boldsymbol{H}^{(L-1)})\boldsymbol{W}^{(L)} + \boldsymbol{\tilde{A}}\boldsymbol{H}^{(1)}\boldsymbol{\hat{W}}^{(L)}\big)
\end{equation}
to keep the properties produced by certain activation (e.g. Sigmoid will produce results in $[0,1]$). Finally, it is important to note that in the VAE framework, the latent variable $\boldsymbol{Z}$ does not naturally fit in the GCN framework where each node is treated as a feature vector. Thus, for the data point-wise distribution version of SAG-VAE, one needs to first transform the dimension into $n$ with a fully-connected layer, and then add one dimension to get a $[m \times n \times 1]$ tensor. In contrast, for the SAG-VAE with dimension-wise distributions, one can directly sample a $[m \times n \times d]$ to operate on GCNs.

\section{Experiments}
\label{sec:experiment}
In this section, we demonstrate the performances of SAG-VAE on various tasks. Intuitively, by learning the graph-structured feature relations, SAG-VAE will have two advantages over ordinary VAEs and their existing variations: interpretable relations and insights between features and robustness against perturbations. To validate the correctness of the learned feature relations, one can apply SAG-VAE to the task of retrieving graph edges based on node feature observations. On the other hand, for the robustness of the SAG-VAE model, one can test the performance on tasks such as reconstruction with noise/mask and sampling with perturbations. \par
For the most of the experiments, the SAG-VAE models are implemented with dimension-wise distributions. The setup is picked for it the advantage of SAG-VAE is more significant with it. The data point-wise distribution counterpart of the SAG-VAE is also straightforward to implement, although the parameters are more difficult to tune. 
\subsection{Graph Connection Retrieval}
We apply two types of feature observations based on graph data. For the first type, the features are generated by a 2-layer Graph Neural Network (GCN in \cite{kipf2016semi}) by propagating information between neighboring nodes; And for the second type, we pick graph data with given feature observations and randomly drop out rows and add Gaussian noises to obtain a collection of noisy data. Notice this task of retrieving graph edge from feature observations is considered as an interesting problem in the area of machine learning. To facilitate the training process, for the SAG-VAE model used for graph connection retrieval, we apply an `informative' prior that adopts the edge density as the prior of Bernoulli distribution. This is a realistic assumption and the type of information is likely available for real-life problems. Thus, it does not affect the fairness of performance comparisons. \par
Results of experiments on two types of graph data illustrate that SAG-VAE can correctly retrieve a significant portion of links (satisfactory recall) while avoid generating overly redundant connections (satisfactory precision). For the first type of data, SAG-VAE can effectively generalize the reconstruction to an unseen pattern of positions. Also, by sampling from the hidden distributions, new patterns of positions can be observed. For the second type of data, SAG-VAE can outperform major existing methods. In addition, the inference of hidden representation is a unique advantage comparing to existing methods. \par
To show the performance advantages of SAG-VAE, the performances of SAG-VAE are compared with pairwise product and Variational Graph Autoencoder (VGAE) \cite{kipf2016variational}. The number of models for comparison is in small scale since there is only limited number of methods capable of inferring links based entirely on feature observations. The most naive model (pairwise product) is to directly compute the dot product between any pair of vertices, and use Sigmoid to produce the probability for a link to exist. This simple method serves as the baseline in the experiments of \cite{kipf2016variational}, although the features in the original experiments were calculated by DeepWalk \cite{perozzi2014deepwalk}. More advanced baselines are based on VGAE, which use part of the graph to learn representation and generalize the generation to the overall graph. The direct comparison between VGAE and SAG-VAE is to remove all edge connections and feed the graph data to VGAE with only `self-loops' on each node. To further validate the superiority of SAG-VAE, we also demonstrate the performance of VGAE with 10\% of edges preserved in the training input, and show that SAG-VAE can outperform VGAE even under this biased setup. 
\subsubsection{Karate Synthetic Data}
We adopt Zachary's karate club graph to generate the first type of feature observations. In the implementation, each type of node (labeled in the original dataset) is parametrized by an individual Gaussian distribution, and 5 different weights are adopted to generate graphs with 5 patterns. During the training phase, only the first 4 types of graphs are provided to the network, and the final pattern is used to test if the trained SAG-VAE is able to generalize the prediction. \par
\begin{figure}[h!]
\centering
\includegraphics[width=0.8\textwidth]{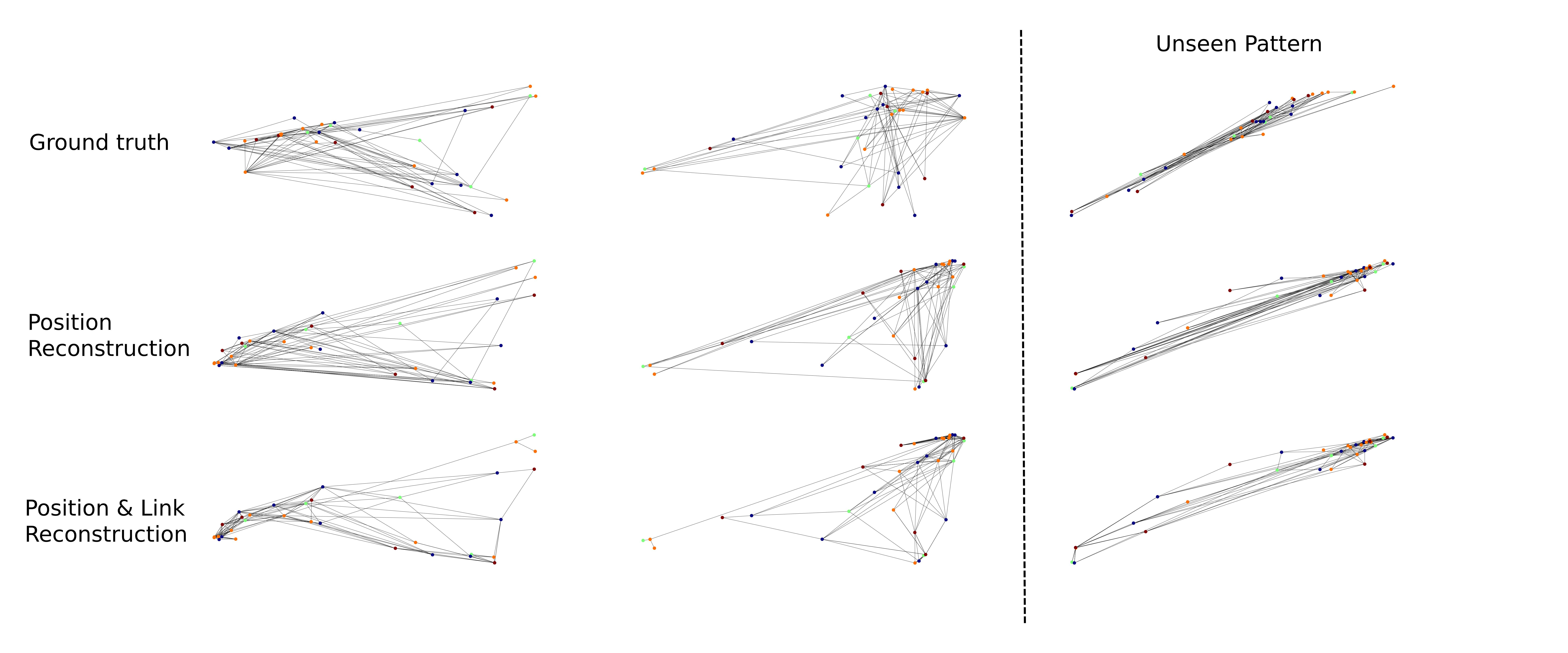}
\caption{\label{fig:KarateReconstruction}SAG-VAE reconstruction of the position and link information of Zachary's karate club data. \textit{Top}: Ground Truth; \textit{Middle}: Position Reconstruction; \textit{Bottom}: Position and Link Reconstruction. Notice that the pattern of the right-most column is not seen by SAG-VAE during the training phase.}
\end{figure}

Figure \ref{fig:KarateReconstruction} illustrates the reconstruction of 3 patterns of node positions based on the SAG-VAE with an individual Gaussian distribution on each dimension. From the figure, it can be observed that the SAG-VAE model can approximately correctly reconstruct the node positions, and while the patterns of links are not exactly the same as the original, the overall geometries are similar in terms of edge distributions. In addition, for the unseen pattern (the rightmost column), the model successfully infers the position and the key links of the graph. 
\begin{figure}[!h]
\centering
\includegraphics[width=0.7\textwidth]{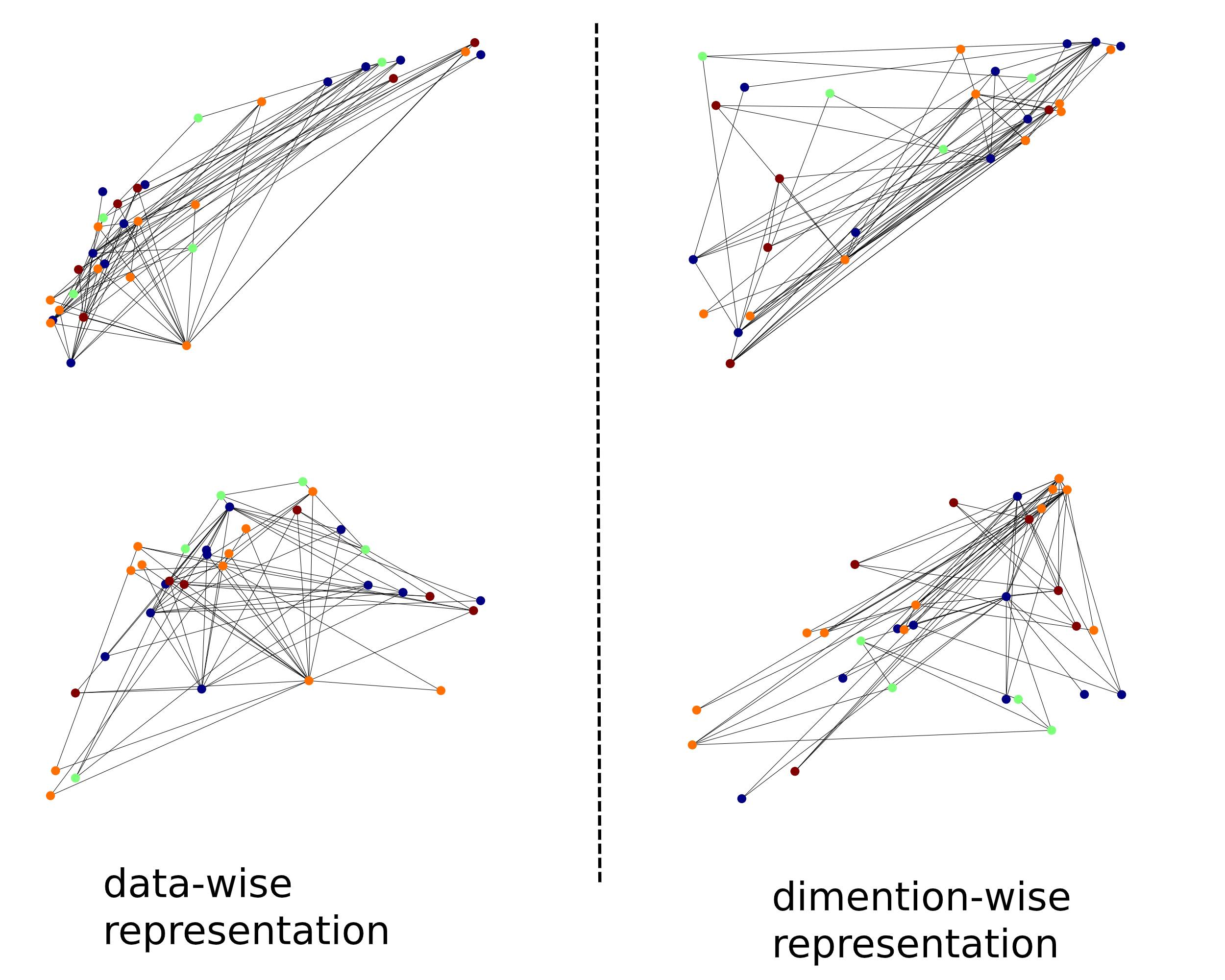}
\caption{\label{fig:karateSampling}Karate position sampling from SAG-VAE with two different implementations}
\end{figure}
\par
Figure \ref{fig:karateSampling} shows the sampling results with both data point- and dimension-wise representation of SAG-VAE. From the figures, it can be observed that both versions of SAG-VAE can generate Karate data information in an organized manner. Sampling from the SAG-VAE with data-wise latent code can further restrict the patterns of the graph, while sampling from its dimension-wise counterpart appears to get a more organized distribution on the node level with different types of nodes better segmented. \par
Table \ref{tab:karate} illustrates the comparison of performance between different methods on the Karate-generated data. From the table it can be observed that SAG-VAE with both data-wise and dimension-wise implementations can outperform methods of comparisons. It is noticeable that for this graph generation task, adding 10\% ground-truth links does not help significantly improve the $F_{1}$ score of VGAE. In contrast, simply applying pairwise product will lead to a better performance in this case.
\begin{table}
\begin{center}
{\caption{Performance comparison between SAG-VAE and other methods on Karate-generated data.}\label{tab:karate}}
\begin{tabular}{p{4cm}|ccc}
Method & Precision & Recall & $F_{1}$ score \\\midrule[1pt]
Pairwise Product & 0.139 & 0.985 & 0.243 \\ \hline
VGAE (no input edge) & 0.142 & 0.524 &  0.223 \\ \hline
VGAE (10 \% link) & 0.150 & 0.539 &  0.234 \\ \hline
SAG-VAE (data-wise) & \textbf{0.616} & \textbf{0.558} &  \textbf{0.586} \\ \hline
SAG-VAE (dimension-wise) & \textbf{0.558} & \textbf{0.611} &  \textbf{0.583} \\ \bottomrule[1pt]
\end{tabular}
\end{center}
\end{table}

\subsubsection{Graph Data with given Node Features}
Table \ref{tab:graphexperiment} illustrates the comparison of performance ($F_{1}$ scores) between different models on three benchmark graph data sets: Graph Protein \cite{borgwardt2005protein,dobson2003distinguishing}, Coil-rag \cite{riesen2008iam,nene1996columbia} and Enzymes \cite{borgwardt2005protein,schomburg2004brenda}. All the 3 types of data come with rich node feature representation, and we obtain the training and testing data by selecting one sub-graph from the data and apply the second type of data generation (with random noise and row dropout). The extracted graph are of size 64, 6 and 18, respectively. Comparing to the Karate data used above, the graphs adopted here are significantly sparser \par
From Table \ref{tab:graphexperiment}, it can be observed that SAG-VAE can outperform methods adopted for comparison, especially for the VGAE-based results. For VGAE, the performance is poor for all datasets and adding back 10\% links does not help remedy the situation. On the other hand, simply applying pairwise product yields in quite competitive performances. One possible reason behind this observation is that since the node features are highly noisy, it is very difficult for the VAE architecture to learn meaningful embedding of the nodes; on the other hand, since the feature representations are originally rich, pairwise product can capture sufficient information, and therefore leads to an unexpected good performance. The curse of noisy feature is resolved by applying SAG-VAE: with the merits of the joint inference of data representation and feature relations, the model can overcome the problem of noise under the VAE framework and lead to overall superior performances.
\begin{table}[h]
\begin{center}
{\caption{Performance comparison ($F_{1}$ score only) between SAG-VAE and other methods on graph data with given node features.}\label{tab:graphexperiment}}
\begin{tabular}{p{4cm}|ccc}
Method & Protein & Colirag & Enzymes \\\midrule[1pt]
Pairwise Product & 0.367 & 0.714 &  0.410\\ \hline
VGAE (no input edge) & 0.276 & 0.620 & 0.315  \\ \hline
VGAE (10 \% link) & 0.283 & 0.643 &  0.319\\ \hline
SAG-VAE (dimension-wise) & \textbf{0.385} & \textbf{0.800} & \textbf{0.423} \\ \bottomrule[1pt]
\end{tabular}
\end{center}
\end{table}

\subsection{Image Data: Robust Reconstruction and Sampling}
\label{sec:expimage}
As it is stated before, we expect SAG-VAE to have a more robust performance against perturbations because of the learned correlations between features can lead to a noise-resisting inductive bias. In this section, we test the robustness of SAG-VAE on two image datasets: MNIST and Fashion MNIST. The performances are evaluated based on 3 tasks: masked/corrupted reconstruction, noisy reconstruction, and noisy sampling. Intuitively, for the reconstruction tasks, if the reconstructed images from SAG-VAE are of higher qualities than those from plain VAE, the robustness of SAG-VAE will be corroborated. Moreover, the noisy sampling task will directly perturb some of the hidden representations, and the inductive bias in SAG-VAE will be able to overcome it. Finally, the plots of the adjacency matrices will show how well the model learned the structured relationships between features. While we may not have any metric to measure it, we can observe if the learned relations are structured and if they are consistent with the characteristics of the images. \par
In these experiments, we only implemented SAG-VAE with dimension-wise distributions. This type of model can produce reconstruction with higher qualities, but it is more vulnerable to perturbation. Therefore, testing with this type of implementation can better illustrate the advantages of SAG-VAE. A drawback of the dimension-wise distribution in sampling is that it makes the data representation harder to obtain, as there is no immediate low-dimension latent codes. Hence, to conduct the sampling process, we model the mean and variance of each pixel for data with different labels. We use Gaussian distribution:
\begin{align*}
\boldsymbol{\mu} \sim \mathcal{N}(\boldsymbol{\mu}_{\mu},\boldsymbol{\sigma}_{\mu}) \qquad \boldsymbol{\sigma} \sim \mathcal{N}(\boldsymbol{\mu}_{\sigma},\boldsymbol{\sigma}_{\sigma})
\end{align*}
to approximately model the manifold and distributions of each dimensions. Notice that unlike graph data, for images, using dimension-wise distribution will bring high image variance. Therefore, it is not recommended to use this strategy in practice. We apply this paradigm here mainly for the purpose to illustrate the robustness of the SAG-VAE.
\subsubsection{Noisy and Masked Reconstruction}
Both MNIST and Fashion MNIST images are in the shape of $28\times28$. For the Fashion MNIST, to better leverage a common structure, we remove the image classes that are not shirt-like or pans-like since their geometries are significantly different from the rest of the dataset. To artificially introduce adversarial perturbation on images, two types of noises are applied: uniform noise and block-masking (corruption). For uniform noise-based perturbation, 200 pixels (150 for Fashion MNIST) are randomly selected and replaced with a number generated from uniform distribution $U(0,1)$. For masked-based perturbation, a block of $6\times6$ is added at random position on each image, thus a small portion of the digit or object in the image is unseen.\par
\begin{figure}[h]
\centering
\includegraphics[width=0.75\textwidth]{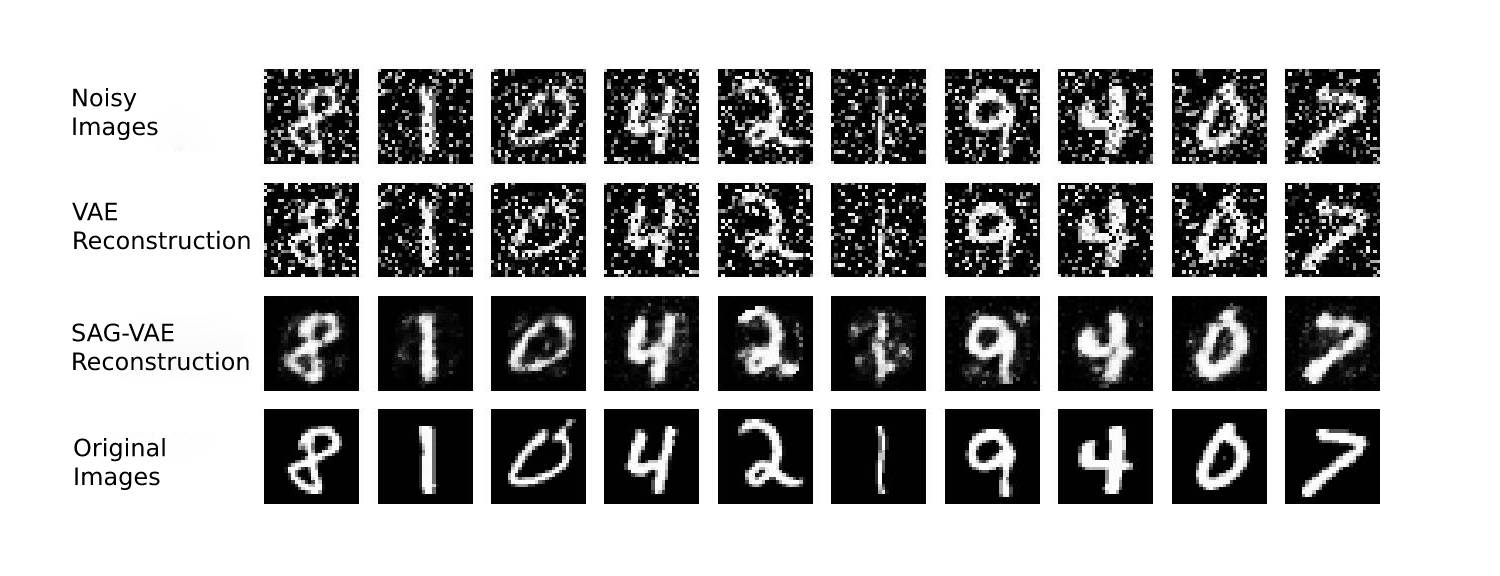}
\caption{\label{fig:VAEs_denoise_mnist}Reconstruction comparison on noisy MNIST. \textit{Top}: Noisy images; \textit{2nd row}: VAE Reconstruction; \textit{3rd row}: SAG-VAE Reconstruction; \textit{Bottom}: Original images.}
\end{figure}
\begin{figure}[h]
\centering
\includegraphics[width=0.75\textwidth]{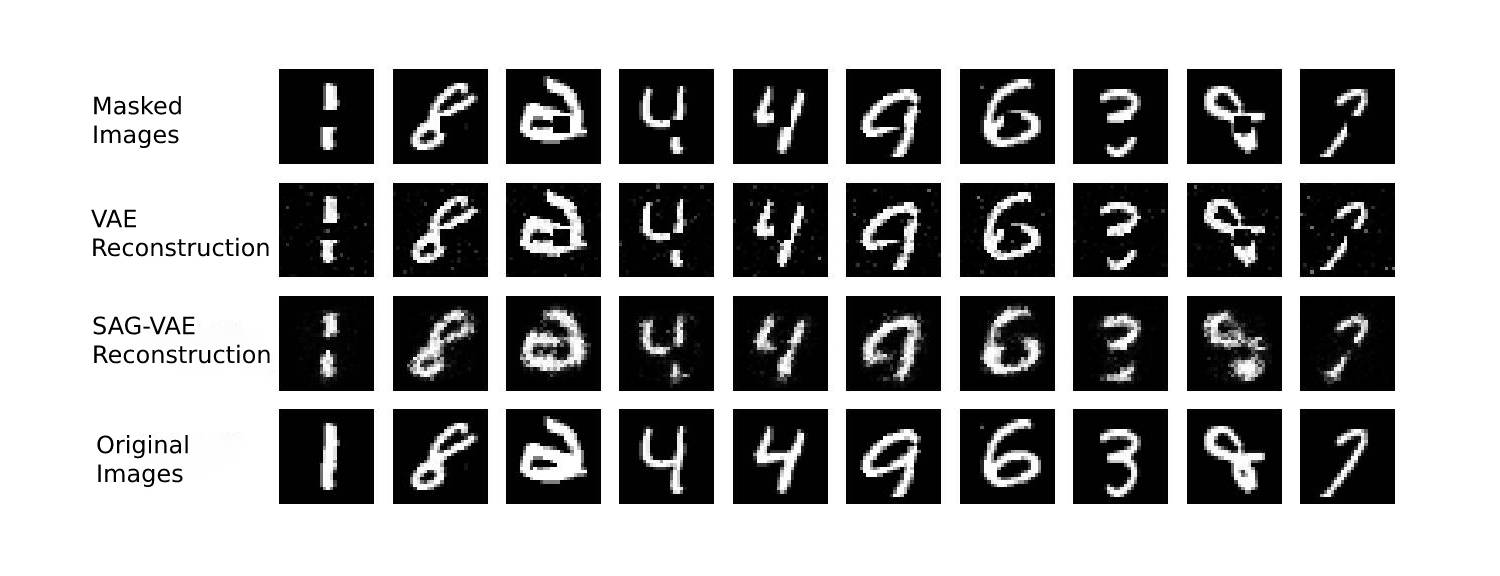}
\caption{\label{fig:VAEs_mask_mnist}Reconstruction comparison on masked MNIST. \textit{Top}: Masked images; \textit{2nd row}: VAE Reconstruction; \textit{3rd row}: SAG-VAE Reconstruction; \textit{Bottom}: Original images.}

\end{figure}

We firstly test SAG-VAE on MNIST data with perturbations. 10 reconstructed images with corresponding perturbed and original images are randomly selected and presented in figure \ref{fig:VAEs_denoise_mnist} and figure \ref{fig:VAEs_mask_mnist}. On the same image, the performance of vanilla VAE is also illustrated. The vanilla VAE is implemented with fully connected layer for each dimension, which is equivalent to SAG-VAE with the adjacency matrix (links) to be zero for all but the main diagonal. \par

As one can observe, images reconstructed by vanilla VAE falsely learned the patterns of noise and blocks, as there is no inductive bias against such situation. On the other hand, for both tasks, SAG-VAE outperforms VAE significantly in terms of reconstruction quality. For the noisy perturbation, one can merely observe visible noise from the reconstruction result of SAG-VAE. And for the masked perturbation, although the reconstruction quality is not as strong, it can still be observed that the edges of blocks are smoothed and mask sizes are reduced adequately. Notice that the performance of SAG-VAE on the task with uniform noise is close to denoising autoencoder \cite{vincent2008extracting}, yet we \emph{did not introduce any explicit denoising measure}. The de-noising characteristics is introduced almost entirely by the inductive bias from the learned feature relations. \par
\begin{figure}[h]
\centering
\includegraphics[width=0.7\textwidth]{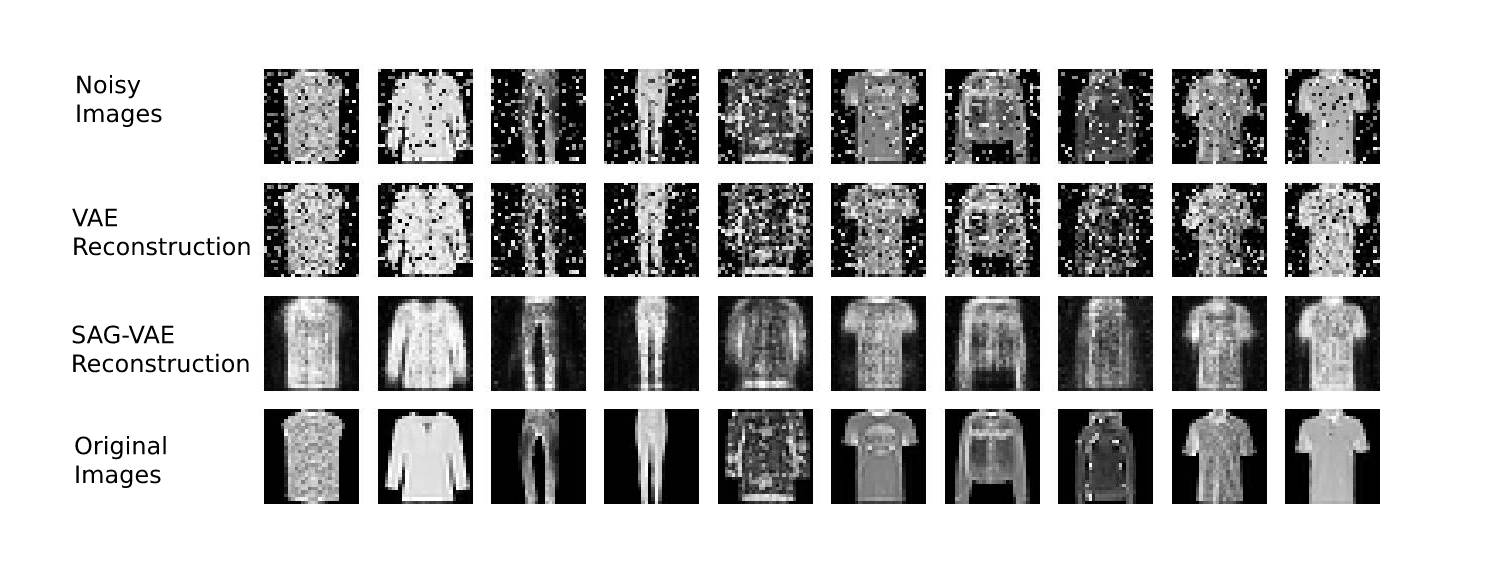}
\caption{\label{fig:VAEs_denoise_fmnist}Reconstruction comparison on noisy Fashion MNIST. \textit{Top}: Noisy images; \textit{2nd row}: VAE Reconstruction; \textit{3rd row}: SAG-VAE Reconstruction; \textit{Bottom}: Original images.}
\end{figure}
\begin{figure}[!h]
\centering
\includegraphics[width=0.7\textwidth]{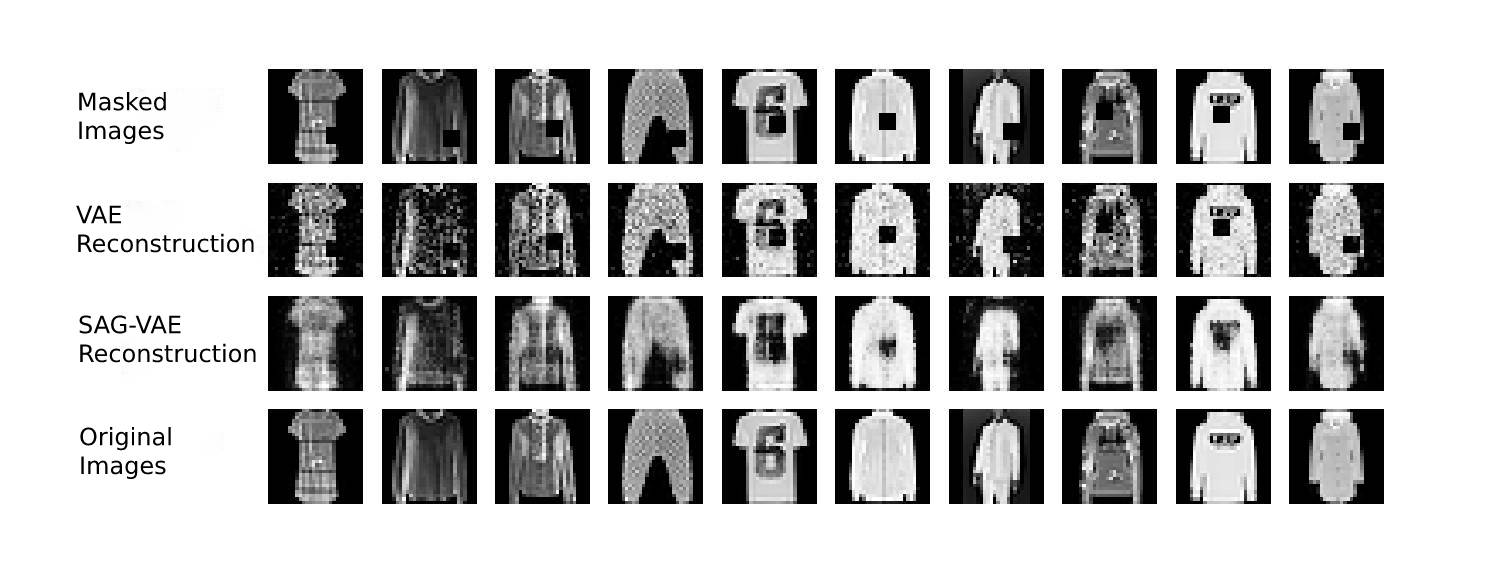}
\caption{\label{fig:VAEs_mask_fmnist}Reconstruction comparison on masked Fashion MNIST. \textit{Top}: Masked images; \textit{2nd row}: VAE Reconstruction; \textit{3rd row}: SAG-VAE Reconstruction; \textit{Bottom}: Original images.}
\end{figure}

We further test the same tasks on Fashion MNIST, and the performances can be shown in figures \ref{fig:VAEs_denoise_fmnist} and \ref{fig:VAEs_mask_fmnist}. Again, we can observe from the figures that SAG-VAE significantly outperforms VAE when perturbation exists in the input data. It is noticeable that in Fashion MNIST reconstruction, SAG-VAE appears to be more resistant to block-masking, although the robustness against uniform noise is much more significant, similar to its performances on the MNIST dataset.\par
\begin{figure}[h]
\centering
\includegraphics[width=0.75\textwidth]{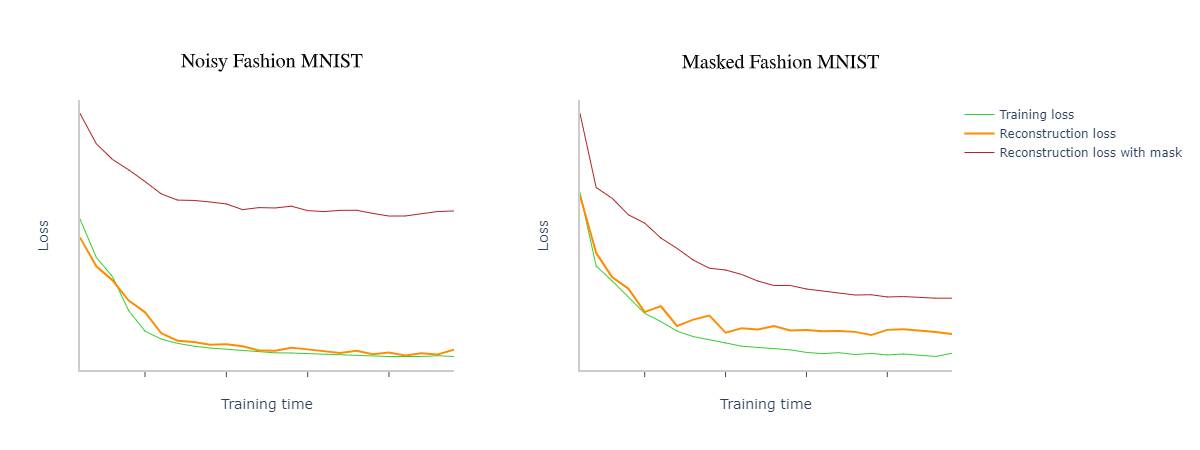}
\caption{\label{fig:fmnist_loss}Training Loss and Reconstruction Loss of Fashion MNIST.}
\end{figure}

Figure \ref{fig:fmnist_loss} shows the loss ($l_{2}$ distance) between reconstructed images and the original and the noise-corrupted images respectively for the SAG-VAE on the Fashion MNIST data. The legends are removed in the interest of the clarity of plotting. It can be observed that the gap between reconstructed and original images declines aligned with training loss, while the loss between reconstructed images and noise images declines ends up with landing at a plateau on a high level. This indicates that the robustness of SAG-VAE will defy itself from learning the perturbation as information. Limited by the space, we did not include the figure for the training losses of vanilla VAE. In our experiments, we observe that for vanilla VAE, the reconstruction loss between the noisy image will continue to decrease while the loss between the real image will increase, indicating that plain VAE falsely fits the perturbation as information. \par
\begin{figure}[!h]
\centering
\includegraphics[width=0.7\textwidth]{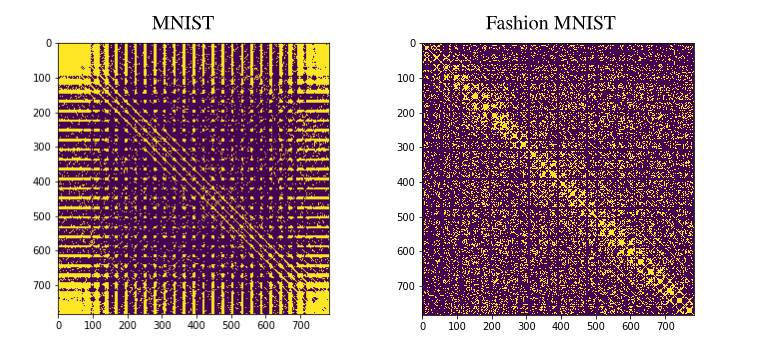}
\caption{\label{fig:ajmtx_mnist_fmnist}Adjacency Matrix Generated from MNIST and Fashion MNIST.}
\end{figure}

Finally, figure \ref{fig:ajmtx_mnist_fmnist} shows the learned feature relations as adjacency matrices for both MNIST and Fashion MNIST. It can be observed that while it is not very straightforward to interpret the reason for each connection to exist, the graph structure is properly organized, and it can be reasonably argues that the robustness against perturbation comes from this organized structure.
\subsubsection{Noisy Sampling}
\label{subsec:noisesample}
\begin{figure}[h]
\centering
\includegraphics[width=0.7\textwidth]{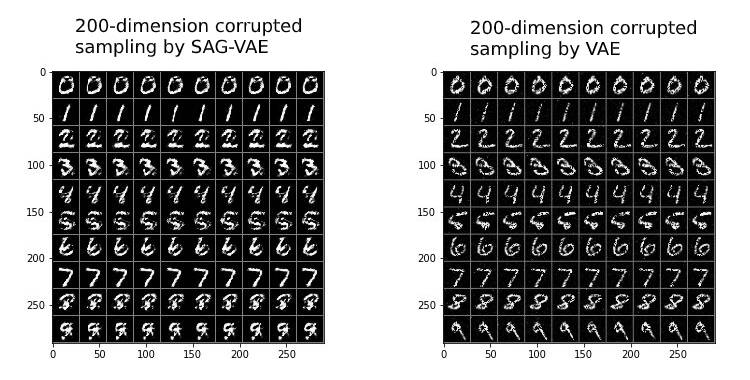}
\caption{\label{fig:sampling}Noisy Sampling on MNIST images.}
\end{figure}

Following the method discussed in section \ref{sec:expimage}, we fit the latent distribution of different MNIST digits/classes with the means and variances of each pixel. For each digit/class, we only pick one image to avoid the high image variance from the dimension-wise modeling. After getting the $\boldsymbol{\mu}$ and $\boldsymbol{\sigma}$ for each digit, we sample $10$ hidden representation $\boldsymbol{z}$ for each of the digits/classes, and randomly replace $200$ dimensions with random noise before sending them to the decoder to generate images. Figure \ref{fig:sampling} illustrates the performance comparison between the SAG-VAE and the vanilla VAE on the above task. \par
From the figure, it can be observed that although both methods can preserve the general manifold of each digit, the SAG-VAE model outperforms vanilla VAE in terms of avoiding `noise over digits', i.e., loss of the grain-like pattern. This can be explained by the graph convolution mechanism of the SAG-VAE, which can `fill' the noise-corrupted pixel through exchanging information with connected pixels. And with a higher quality of image coherence, we can argue that the SAG-VAE model is shown to be more robust against noise perturbations during sampling.

\section{Conclusion}
\label{sec:conclusion}
In this paper, we propose Self-Attention Graph Variaional AutoEncoder (SAG-VAE) based on recent advances on Variational Autoencoders and Graph Neural Networks. This novel model can jointly infer data representations and relations between features, which provides strong explainable results for the input datasets. In addition, by introducing the learned relations as inductive biases, the model demonstrates strong robustness against perturbations. Besides, a novel Self-Attention Graph Neural Network (SA-GNN) is proposed in the paper. \par
To conclude, this paper makes the following major contributions: firstly, it proposes a novel VAE-based framework which can jointly infer representations and feature relations in an end-to-end manner; secondly, it presents a novel Self-attention-based Graph Neural Network, which leverages the power of self-attention mechanism to improve the performance; and finally, it demonstrates advantageous performances on multiple experiments, which can be of great utility in practice. \par
In the future, the authors intend to extend the model to more advanced posterior approximation techniques (e.g. IWAE) and more flexible priors (e.g. normalized flow). Testing the performances of the model on more complicated datasets is another direction.

\bibliographystyle{unsrt}  
\bibliography{references}

\begin{thebibliography}{10}

\bibitem{bordes2013translating}
Antoine Bordes, Nicolas Usunier, Alberto Garcia-Duran, Jason Weston, and Oksana
  Yakhnenko.
\newblock Translating embeddings for modeling multi-relational data.
\newblock In {\em Advances in neural information processing systems}, pages
  2787--2795, 2013.

\bibitem{xu2018representation}
Keyulu Xu, Chengtao Li, Yonglong Tian, Tomohiro Sonobe, Ken ichi Kawarabayashi,
  and Stefanie Jegelka.
\newblock Representation learning on graphs with jumping knowledge networks.
\newblock In {\em International Conference on Machine Learning}, 2018.

\bibitem{van2018relational}
Sjoerd Van~Steenkiste, Michael Chang, Klaus Greff, and J{\"u}rgen Schmidhuber.
\newblock Relational neural expectation maximization: Unsupervised discovery of
  objects and their interactions.
\newblock In {\em International Conference on Learning Representations}, 2018.

\bibitem{battaglia2018relational}
Peter~W Battaglia, Jessica~B Hamrick, Victor Bapst, Alvaro Sanchez-Gonzalez,
  Vinicius Zambaldi, Mateusz Malinowski, Andrea Tacchetti, David Raposo, Adam
  Santoro, Ryan Faulkner, et~al.
\newblock Relational inductive biases, deep learning, and graph networks.
\newblock {\em arXiv preprint arXiv:1806.01261}, 2018.

\bibitem{louizos2017causal}
Christos Louizos, Uri Shalit, Joris~M Mooij, David Sontag, Richard Zemel, and
  Max Welling.
\newblock Causal effect inference with deep latent-variable models.
\newblock In {\em Advances in Neural Information Processing Systems}, pages
  6446--6456, 2017.

\bibitem{kipf2018neural}
Thomas Kipf, Ethan Fetaya, Kuan-Chieh Wang, Max Welling, and Richard Zemel.
\newblock Neural relational inference for interacting systems.
\newblock In {\em International Conference on Machine Learning}, 2018.

\bibitem{sanchez2018graph}
Alvaro Sanchez-Gonzalez, Nicolas Heess, Jost~Tobias Springenberg, Josh Merel,
  Martin Riedmiller, Raia Hadsell, and Peter Battaglia.
\newblock Graph networks as learnable physics engines for inference and
  control.
\newblock In {\em International Conference on Machine Learning}, 2018.

\bibitem{scarselli2008graph}
Franco Scarselli, Marco Gori, Ah~Chung Tsoi, Markus Hagenbuchner, and Gabriele
  Monfardini.
\newblock The graph neural network model.
\newblock {\em IEEE Transactions on Neural Networks}, 20(1):61--80, 2008.

\bibitem{kipf2016semi}
Thomas~N Kipf and Max Welling.
\newblock Semi-supervised classification with graph convolutional networks.
\newblock In {\em International Conference on Learning Representations}, 2017.

\bibitem{wu2019comprehensive}
Zonghan Wu, Shirui Pan, Fengwen Chen, Guodong Long, Chengqi Zhang, and Philip~S
  Yu.
\newblock A comprehensive survey on graph neural networks.
\newblock {\em IEEE Transactions on Neural Networks and Learning Systems},
  2020.

\bibitem{kingma2013auto}
Diederik~P Kingma and Max Welling.
\newblock Auto-encoding variational bayes.
\newblock In {\em International Conference on Learning Representations}, 2014.

\bibitem{schlichtkrull2018modeling}
Michael Schlichtkrull, Thomas~N Kipf, Peter Bloem, Rianne Van Den~Berg, Ivan
  Titov, and Max Welling.
\newblock Modeling relational data with graph convolutional networks.
\newblock In {\em European Semantic Web Conference}, pages 593--607. Springer,
  2018.

\bibitem{jang2016categorical}
Eric Jang, Shixiang Gu, and Ben Poole.
\newblock Categorical reparameterization with gumbel-softmax.
\newblock In {\em International Conference on Learning Representations}, 2017.

\bibitem{granger1969investigating}
Clive~WJ Granger.
\newblock Investigating causal relations by econometric models and
  cross-spectral methods.
\newblock {\em Econometrica: Journal of the Econometric Society}, pages
  424--438, 1969.

\bibitem{kuipers1984causal}
Benjamin Kuipers and Jerome~P Kassirer.
\newblock Causal reasoning in medicine: analysis of a protocol.
\newblock {\em Cognitive Science}, 8(4):363--385, 1984.

\bibitem{brillinger1976identification}
David~R Brillinger, Hugh~L Bryant, and Jose~P Segundo.
\newblock Identification of synaptic interactions.
\newblock {\em Biological cybernetics}, 22(4):213--228, 1976.

\bibitem{watts1998collective}
Duncan~J Watts and Steven~H Strogatz.
\newblock Collective dynamics of ‘small-world’networks.
\newblock {\em nature}, 393(6684):440, 1998.

\bibitem{linderman2016bayesian}
Scott Linderman, Ryan~P Adams, and Jonathan~W Pillow.
\newblock Bayesian latent structure discovery from multi-neuron recordings.
\newblock In {\em Advances in neural information processing systems}, pages
  2002--2010, 2016.

\bibitem{yang2018glomo}
Zhilin Yang, Jake Zhao, Bhuwan Dhingra, Kaiming He, William~W Cohen, Russ~R
  Salakhutdinov, and Yann LeCun.
\newblock Glomo: unsupervised learning of transferable relational graphs.
\newblock In {\em Advances in Neural Information Processing Systems}, pages
  8950--8961, 2018.

\bibitem{hochreiter1997long}
Sepp Hochreiter and J{\"u}rgen Schmidhuber.
\newblock Long short-term memory.
\newblock {\em Neural computation}, 9(8):1735--1780, 1997.

\bibitem{mikolov2013efficient}
Tomas Mikolov, Kai Chen, Greg Corrado, and Jeffrey Dean.
\newblock Efficient estimation of word representations in vector space.
\newblock In {\em International Conference on Learning Representations}, 2013.

\bibitem{eslami2016attend}
SM~Ali Eslami, Nicolas Heess, Theophane Weber, Yuval Tassa, David Szepesvari,
  Geoffrey~E Hinton, et~al.
\newblock Attend, infer, repeat: Fast scene understanding with generative
  models.
\newblock In {\em Advances in Neural Information Processing Systems}, pages
  3225--3233, 2016.

\bibitem{greff2017neural}
Klaus Greff, Sjoerd van Steenkiste, and J{\"u}rgen Schmidhuber.
\newblock Neural expectation maximization.
\newblock In {\em Advances in Neural Information Processing Systems}, pages
  6691--6701, 2017.

\bibitem{sperduti1997supervised}
Alessandro Sperduti and Antonina Starita.
\newblock Supervised neural networks for the classification of structures.
\newblock {\em IEEE Transactions on Neural Networks}, 8(3):714--735, 1997.

\bibitem{gori2005new}
Marco Gori, Gabriele Monfardini, and Franco Scarselli.
\newblock A new model for learning in graph domains.
\newblock In {\em Proceedings. 2005 IEEE International Joint Conference on
  Neural Networks, 2005.}, volume~2, pages 729--734. IEEE, 2005.

\bibitem{bruna2013spectral}
Joan Bruna, Wojciech Zaremba, Arthur Szlam, and Yann LeCun.
\newblock Spectral networks and locally connected networks on graphs.
\newblock In {\em International Conference on Learning Representations}, 2014.

\bibitem{defferrard2016convolutional}
Micha{\"e}l Defferrard, Xavier Bresson, and Pierre Vandergheynst.
\newblock Convolutional neural networks on graphs with fast localized spectral
  filtering.
\newblock In {\em Advances in neural information processing systems}, pages
  3844--3852, 2016.

\bibitem{henaff2015deep}
Mikael Henaff, Joan Bruna, and Yann LeCun.
\newblock Deep convolutional networks on graph-structured data.
\newblock {\em arXiv preprint arXiv:1506.05163}, 2015.

\bibitem{Wang2018NerveNetLS}
Tingwu Wang, Renjie Liao, Jimmy Ba, and Sanja Fidler.
\newblock Nervenet: Learning structured policy with graph neural networks.
\newblock In {\em International Conference on Learning Representations}, 2018.

\bibitem{Grover2019GraphiteIG}
Aditya Grover, Aaron Zweig, and Stefano Ermon.
\newblock Graphite: Iterative generative modeling of graphs.
\newblock In {\em International Conference on Machine Learning}, 2019.

\bibitem{velivckovic2017graph}
Petar Veli{\v{c}}kovi{\'c}, Guillem Cucurull, Arantxa Casanova, Adriana Romero,
  Pietro Lio, and Yoshua Bengio.
\newblock Graph attention networks.
\newblock In {\em International Conference on Learning Representations}, 2018.

\bibitem{kipf2016variational}
Thomas~N Kipf and Max Welling.
\newblock Variational graph auto-encoders.
\newblock {\em arXiv preprint arXiv:1611.07308}, 2016.

\bibitem{pan2018adversarially}
Shirui Pan, Ruiqi Hu, Guodong Long, Jing Jiang, Lina Yao, and Chengqi Zhang.
\newblock Adversarially regularized graph autoencoder for graph embedding.
\newblock In {\em International Joint Conference on Artificial Intelligence},
  2018.

\bibitem{chou2019generated}
Jason Chou.
\newblock Generated loss and augmented training of mnist vae.
\newblock {\em arXiv preprint arXiv:1904.10937}, 2019.

\bibitem{zhang2018self}
Han Zhang, Ian Goodfellow, Dimitris Metaxas, and Augustus Odena.
\newblock Self-attention generative adversarial networks.
\newblock In {\em International Conference on Machine Learning}, 2019.

\bibitem{dieng2018avoiding}
Adji~B Dieng, Yoon Kim, Alexander~M Rush, and David~M Blei.
\newblock Avoiding latent variable collapse with generative skip models.
\newblock In {\em AISTATS}, 2019.

\bibitem{perozzi2014deepwalk}
Bryan Perozzi, Rami Al-Rfou, and Steven Skiena.
\newblock Deepwalk: Online learning of social representations.
\newblock In {\em Proceedings of the 20th ACM SIGKDD international conference
  on Knowledge discovery and data mining}, pages 701--710. ACM, 2014.

\bibitem{borgwardt2005protein}
Karsten~M Borgwardt, Cheng~Soon Ong, Stefan Sch{\"o}nauer, SVN Vishwanathan,
  Alex~J Smola, and Hans-Peter Kriegel.
\newblock Protein function prediction via graph kernels.
\newblock {\em Bioinformatics}, 21(suppl\_1):i47--i56, 2005.

\bibitem{dobson2003distinguishing}
Paul~D Dobson and Andrew~J Doig.
\newblock Distinguishing enzyme structures from non-enzymes without alignments.
\newblock {\em Journal of molecular biology}, 330(4):771--783, 2003.

\bibitem{riesen2008iam}
Kaspar Riesen and Horst Bunke.
\newblock Iam graph database repository for graph based pattern recognition and
  machine learning.
\newblock In {\em Joint IAPR International Workshops on Statistical Techniques
  in Pattern Recognition (SPR) and Structural and Syntactic Pattern Recognition
  (SSPR)}, pages 287--297. Springer, 2008.

\bibitem{nene1996columbia}
Sameer~A Nene, Shree~K Nayar, Hiroshi Murase, et~al.
\newblock Columbia object image library (coil-20).
\newblock 1996.

\bibitem{schomburg2004brenda}
Ida Schomburg, Antje Chang, Christian Ebeling, Marion Gremse, Christian Heldt,
  Gregor Huhn, and Dietmar Schomburg.
\newblock Brenda, the enzyme database: updates and major new developments.
\newblock {\em Nucleic acids research}, 32(suppl\_1):D431--D433, 2004.

\bibitem{vincent2008extracting}
Pascal Vincent, Hugo Larochelle, Yoshua Bengio, and Pierre-Antoine Manzagol.
\newblock Extracting and composing robust features with denoising autoencoders.
\newblock In {\em Proceedings of the 25th international conference on Machine
  learning}, pages 1096--1103. ACM, 2008.

\end{thebibliography}

\end{document}